\documentclass[10pt, a4paper]{article}
\usepackage{amsmath}
\usepackage{CJKutf8}
\usepackage{lrec-coling2024} 
\title{Restoring Ancient Ideograph: A Multimodal Multitask \\Neural Network Approach}

\name{Siyu Duan$^{1,3}$, Jun Wang$^{1,3,4}$,Qi Su$^{2,3,4}$\sthanks{Corresponding Author: Qi Su, sukia@pku.edu.cn}}
\address{
    $^{1}$ Department of Information Management, Peking University\\
    $^{2}$School of Foreign Languages, Peking University \\
    $^{3}$Center for Digital Humanities, Peking University \\
    $^{4}$Institute for Artificial Intelligence, Peking University \\
    \texttt{\{duansiyu, junwang, sukia\}@pku.edu.cn}
    }

\abstract{
Cultural heritage serves as the enduring record of human thought and history. Despite significant efforts dedicated to the preservation of cultural relics, many ancient artefacts have been ravaged irreversibly by natural deterioration and human actions. Deep learning technology has emerged as a valuable tool for restoring various kinds of cultural heritages, including ancient text restoration. Previous research has approached ancient text restoration from either visual or textual perspectives, often overlooking the potential of synergizing multimodal information. This paper proposes a novel Multimodal Multitask Restoring Model (MMRM) to restore ancient texts, particularly emphasising the ideograph. This model combines context understanding with residual visual information from damaged ancient artefacts, enabling it to predict damaged characters and generate restored images simultaneously. We tested the MMRM model through experiments conducted on both simulated datasets and authentic ancient inscriptions. The results show that the proposed method gives insightful restoration suggestions in both simulation experiments and real-world scenarios. To the best of our knowledge, this work represents the pioneering application of multimodal deep learning in ancient text restoration, which will contribute to the understanding of ancient society and culture in digital humanities fields.
\\ \newline  }
\begin{document}
\begin{CJK*}{UTF8}{gbsn}
\maketitleabstract

\section{Introduction}
Ancient cultural heritage stands as a record of human civilization, aiding in the understanding of human history and culture. Regrettably, many ancient artefacts have fallen prey to the ravages of time, natural deterioration, or deliberate human actions, expecting preservation and restoration. Deep learning technology has witnessed a series of remarkable advancements in the restoration of ancient cultural relics, including pottery \citep{nacer2021,cecilia2020}, architecture \citep{zou2021}, murals \citep{wang2018,zeng2020}, etc. Among the myriad facets of cultural heritage, written language is the quintessential vessel of human thought, recording human history with symbols. Restoring ancient texts aimed at proffering suggestions for the attribution of the fragmented scripts. Conventional methods for this task have leaned upon the knowledge of domain experts and the meticulous investigation of literature, which requires the mastery of philology and linguistics, rendering this undertaking a formidable and specialized task.

In this work, we applied the multimodal deep learning methodology to restore ancient texts, with a particular emphasis on the ideograph. Ideograms encapsulate semantics within visual symbols and endow each character with an intuitive visual correspondence. Consequently, restoring the ancient ideogram hinges on contextual information and visual cues. In this paper, we propose a novel Multimodal Multitask Restoring Model (MMRM) for ideograph restoration, synthesising cognizable context and the residual visual message of the damaged artefact to attribute damaged characters. It also employs a multitask learning paradigm to predict the damaged characters and generate restored images simultaneously.

We tested the MMRM model by experiments on both simulated data and authentic ancient inscriptions. The simulation experiments evince that the proposed MMRM model brings a substantive enhancement in ideograph restoration. In real-world scenarios, the model trained in the simulation experiment demonstrates its capacity to provide judicious recommendations for damaged characters. To the best of our knowledge, this work represents the pioneering application of multimodal deep learning methods to the restoration of ancient texts. By contributing to the building and improvement of ancient corpora, which is the basis of numerous digital humanities studies \citep{duan2023disentangling, wang2024evol}, this work will benefit historical, literary, and archaeological scholarship in the contemporary digital milieu.

\section{Related Works}
Ancient text restoration can be approached from two aspects: visual and textual. Visual-based restoration methods involved two computer vision techniques: handwritten text recognition and image inpainting. Ancient handwritten text recognition was extensively studied \citep{narang2020ancient}, however, this is generally for undamaged characters. Image-inpainting-based text restoration endeavours to regenerate the image of damaged texts. In character-level scenarios, the focus is the regeneration of deteriorated strokes. For example, \citet{SU2022} and \citet{chen2022dual} use GAN to reconstruct ancient Chinese Han and Yi texts, respectively. Since the damage in this context often does not impede text recognition, the restoration purpose is to acquire higher-quality images. In document-level scenarios, the emphasis gravitates towards the amelioration of damaged areas, such as ink seepage, watermarks, blemishes, and blur \citep{Wadhwani2021, souibgui2022}. These works aspire to procure higher-fidelity document images, not constricted to the meticulous reconstruction of individual words or characters.

Predicting blank spaces within sentences is a familiar task in natural language processing \citep{shen-etal-2020-blank, donahue-etal-2020-enabling}, which has been applied in the preliminary attempted work of ancient text restoration \citep{assael-etal-2019-restoring, ethan2020}. The pre-trained models also benefit to ancient text restoration since the Masked Language Model (MaskLM) task is widely employed \citep{devlin-etal-2019-bert, liu2019roberta}. For instance, \citet{lazar-etal-2021-filling} introduced multilingual pre-training for Ancient Akkadian text restoration. However, low-resource languages lack sufficient data for pre-training, thus the rudimentary RNN still needs attention, a salient instance being the Mycenaean Greek \citep{Papavassileiou2023}, where training samples are exceedingly scarce. Overall, these methods construct training data by masking out parts of characters in ancient texts, without utilizing visual signals.

Given the paucity of exploitable information within ancient texts, multitask learning fortified by supplementary guidance data was applied in text restoration tasks in some works. For example, the ancillary tasks predicting the region and age of text elicited a discernible enhancement in Ancient Greek restoration \citep{assael2022restoring}. However, annotating ancient texts with their provenance and temporal message constitutes a labour-intensive undertaking. \citet{kang-etal-2021-restoring} employed the simultaneous pursuit of machine translation and text restoration tasks, while its enhancements in text restoration accuracy remained elusive. It is imperative to underscore that these multitask learning methods necessitate supplementary data annotations or parallel corpora. Besides, although both visual and contextual information play important roles in text restoration, the application of multimodal methodology is still a blank area.

\section{Method}
\subsection{Task Define}
This article proposes a multimodal ideograph restoration task that utilizes both visual and textual cues to predict damaged characters in ancient texts. The visual cue is the image of the damaged characters from cultural relics, and the textual cue is the recognizable context text. To restore the damaged ideogram, the model takes the undamaged context and the damaged image as input and predicts the damaged character.
\subsection{Data}

Ancient Chinese is a kind of widely used ideograph. It evolved from hieroglyphs and retained the characteristic of using visual elements to convey semantic meaning. Our experimental data consists of three parts: Classical Chinese corpus to simulate damaged context, images to simulate individual damaged characters and real inscription data to examine the model in a real-world scenario. \footnote{Data and Code: \url{https://github.com/CissyDuan/MMRM}.} In this section, we will introduce how to simulate damaged text and images in our experiments.
\subsubsection{Damaged Text}
The ancient literature data was collected from the publicly available website `xueheng (\url{http://core.xueheng.net/})'. This Classical Chinese corpus includes the core classics in Chinese history, spanning over 2000 years with diverse topics. We segmented the texts into sentences, and when a sentence exceeded 50 characters, it was split into two. We got approximately 590,000 sentences for the simulation experiments. Its statistical information is shown in Table \ref{datastat}.

\begin{table}[!h]
\begin{center}
\begin{tabularx}{\columnwidth}{|l|l|l|X|X|}
      \hline
      Train&Dev&Test&Max&Avg\\
      \hline
      575,398&10,000&10,000&50&14.4\\
      \hline
\end{tabularx}
\caption{Statistics of textual dataset}
\label{datastat}
 \end{center}
\end{table}

In the simulation experiments, the characters in each sentence were randomly masked. It is worth noting that Classical Chinese was written in traditional Chinese characters, while in modern mainland China, simplified Chinese characters are used. This has led to the fact that most common Chinese pre-trained models support simplified characters. Therefore, for each sentence, we prepared both simplified and traditional versions \footnote{The conversion tool is OpenCC (\url{https://github.com/BYVoid/OpenCC})}, one for inputting into the model and one for generating images of characters.

\begin{figure}[t]
\begin{center}
\includegraphics[width=0.99\linewidth]{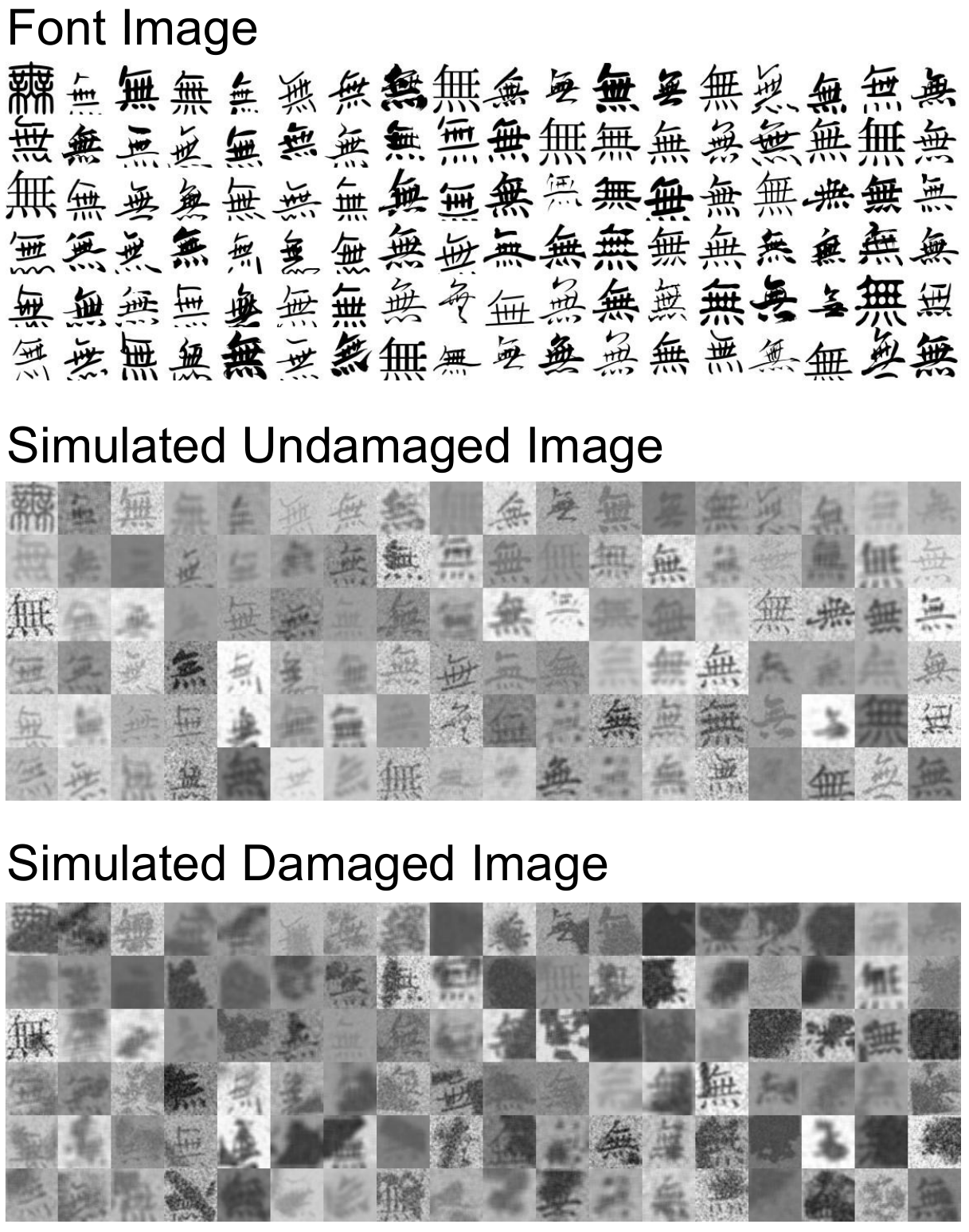} 
\caption{The process of simulating images of damaged characters. The first picture shows images of ideograms generated by 108 kinds of fonts. The second picture shows the simulated images of undamaged characters. The third picture shows the simulated images of damaged characters.}
\label{data}
\end{center}
\end{figure}

\subsubsection{Damaged Image}
After obtaining the text with random blanks, we proceeded to generate simulated images for the damaged characters. In the simulation experiments, we transformed all images to grayscale mode and let the dark text be generated on a light background. This means that in scenarios with light-coloured text on a dark background, such as rubbings of engraved inscriptions, colour inversion needs to be applied in advance.

The simulation of the damaged image involves three steps: 
First, we generate binary images of the characters with random fonts. 
Next, various image processing techniques are applied to simulate the images of undamaged characters collected from cultural relics. 
Finally, by adding random masks to the images, we simulate the images of damaged characters. 
Some simulated samples are shown in the Figure \ref{data}.

\textbf{Step 1}: Generating Font Image

Various calligraphic styles should be considered to simulate images of ancient characters. We have collected 108 traditional Chinese fonts from the internet to simulate diverse calligraphic styles. Images for missing characters will be generated with random fonts. 

\textbf{Step 2}: Simulating Undamaged Images

Each specific font image for a character is unique, which obviously cannot account for text images collected in real-world scenarios. To achieve this, additional image processing is applied, including the random applications of texture, brightness, contrast, translations, rotations, scaling, and Gaussian blur.


\begin{figure}[t]
\begin{center}
\includegraphics[width=0.8\linewidth]{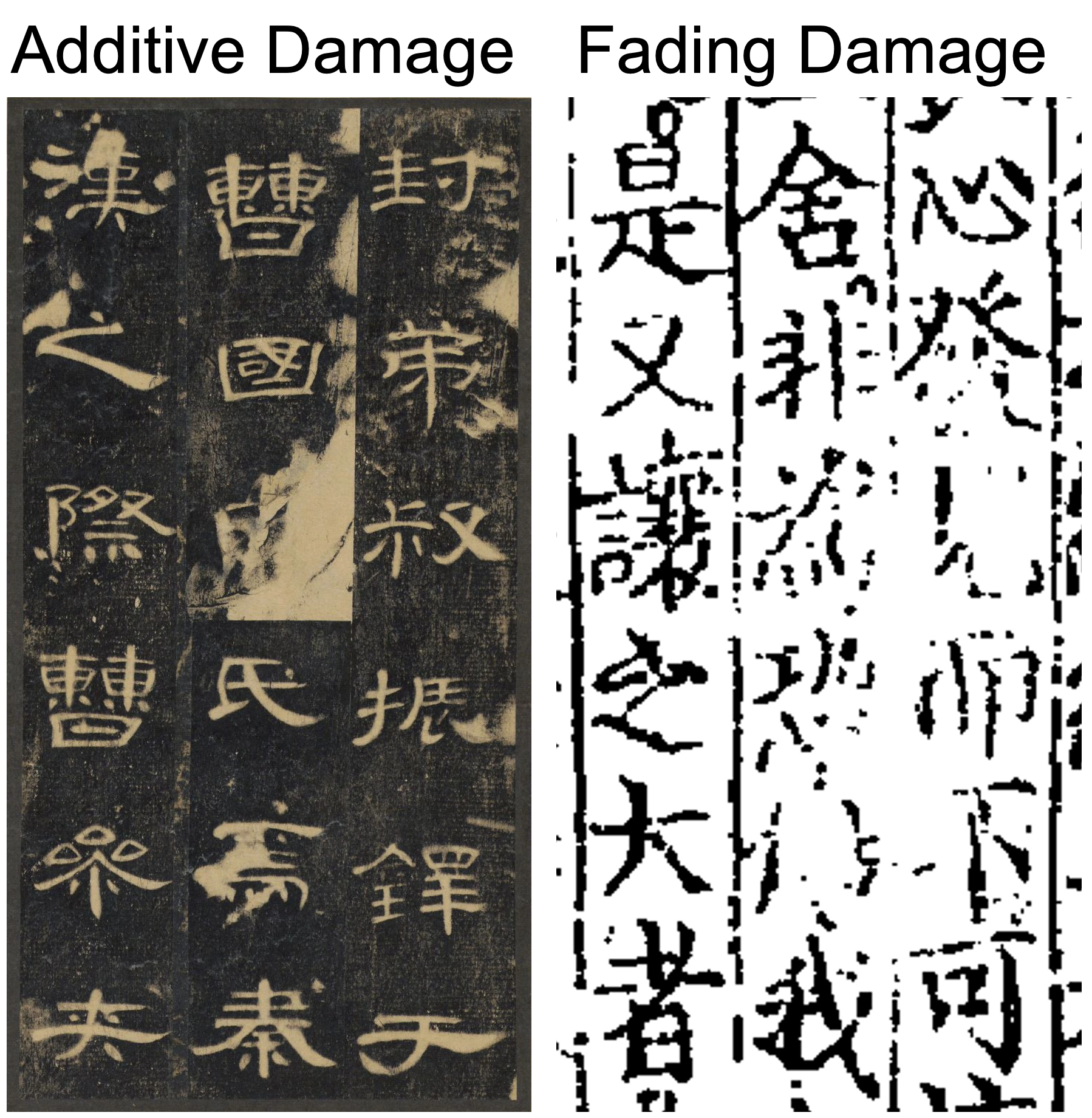} 
\caption{Two concrete cases of Additive Damage and Fading Damage. The left image is from a rubbing of the `Cao Quan Stele'; the right image is from the paper book `\textit{Sima Fa Ji Jie}'.}
\label{damagetype}
\end{center}
\end{figure}

\textbf{Step 3}: Simulating Damaged Images

Two types of noises typically occur on the images of damaged texts: one that is close in colour to the text itself, and the other close in colour to the background. We refer to them as `additive damage' and `fading damage' respectively. We showed two concrete cases in Figure \ref{damagetype}. In these two cases, additive damage is more predominant in the left engraved inscription rubbing, while fading damage is more noticeable in the right paper book. To simulate these damages, we randomly add large masks as well as a random number of spot-type masks. In practical applications, the types, areas, and quantities of masks can be adjusted based on the characteristics of the real conditions. 



\subsection{Model}
\begin{figure}[t]
\begin{center}
\includegraphics[width=0.99\linewidth]{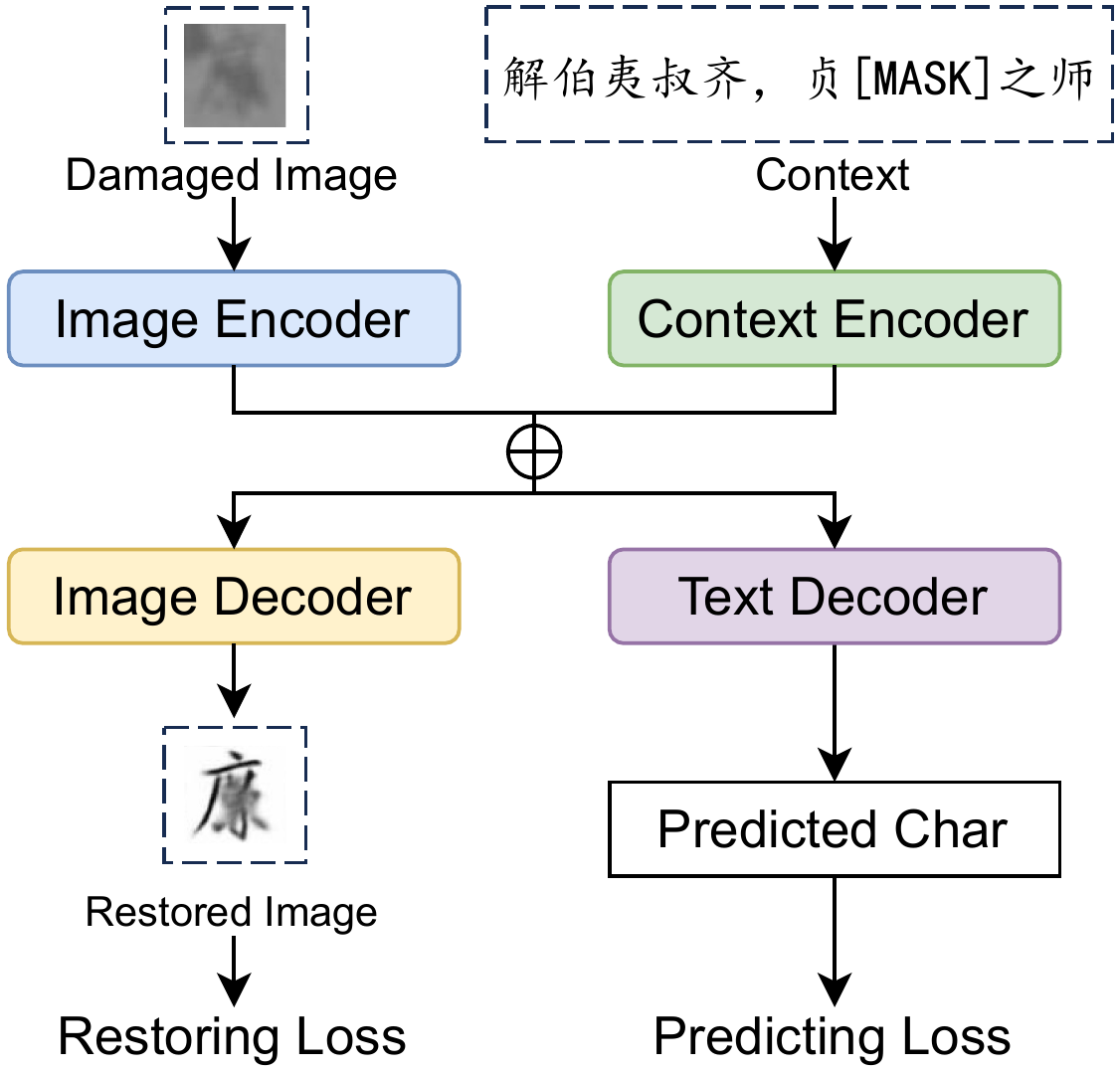} 
\caption{The formulation of Multimodal Multitask Restoration Model (MMRM).}
\label{model}
\end{center}
\end{figure}

We propose a multimodal multitask framework for ideograph restoration, employing four modules to encode and decode the text and image respectively. The model finally gives two types of outputs: the candidate character and the restored image. The model formulation is shown in Figure \ref{model}.

\subsubsection{Modules}
\textbf{Context Encoder}. The context encoder is used to extract contextual features of the damaged text. We employ a pre-trained $RoBERTa$ model to encode the masked context and extract the feature vectors of the masked positions. Before the multimodal training, we fine-tuned the $RoBERTa$ model on the textual dataset by predicting the missing characters based on context alone. In the subsequent multimodal training, these parameters are frozen. The masked position is denoted as $i$, the calculation of this module is as follows:
\begin{align}
    memory&=RoBERTa(Context)\\
    x_1&=memory[i]
\end{align}

\textbf{Image Encoder}. The image encoder is used to extract visual features from the damaged image. We apply a pre-trained $ResNet\ 50$ model and replace the final layer with a new linear layer that maps the features to the same dimension as the Context Encoder. Since the Context Encoder, which has undergone pre-training and fine-tuning, already has preliminary capabilities in predicting the missing character, this linear layer is initialized with all zeros to gradually increase the influence of visual features.
\begin{align}
    x_2=Resnet50(Img)
\end{align}

\textbf{Feature Fusion}. We use additive fusion to combine contextual and visual features. The fused features encompass information from both the text and the image, which will be used to predict the damaged characters and restore the damaged images.
\begin{align}
    x=x_1+x_2
\end{align}
Since the pre-trained and finetuned textual features already possess preliminary capabilities for this task, the image features need to be learned anew. The use of additive fusion, coupled with the full zero initialization of the last linear layer in the image encoder, minimizes interference from image features on textual features during the early stages of training. 

\textbf{Text Decoder}. The fused features will pass through an MLP layer to predict the missing character. This MLP layer is initialized with parameters from the $RoBERTa$ LM layer.
\begin{align}
    Y_{pred}=MLP(x)
\end{align}

\textbf{Image Decoder}. The fused features will pass through multiple transposed convolution layers to generate the restored image.
\begin{align}
    Img_{res}=ConvT(x)
\end{align}
After these processes, the candidate characters and images are generated by the model.

\subsubsection{Multitask Learning}
Highly correlated multitask learning can leverage information sharing between different tasks to improve their respective performances. In the proposed model, we have it not only predict missing characters but also generate images of the restored characters. The multitask learning serves two purposes: one is to predict the missing characters, and the other is to make the restored image resemble the original font image. This is achieved by optimizing two loss functions. One is the restoring loss, which is the MSE loss between the restored image and the font image:
\begin{align}
    Loss_{res}=MSELoss(Img_{res},Img_{font})
\end{align}
One is the predicting loss, which is the cross-entropy loss between the predicted token and the actual character:
\begin{align}
    Loss_{pred}=CELoss(Y_{pred},Label)
\end{align}
The final loss is the weighted sum of the restoring loss and the predicting loss:
\begin{align}
    Loss=\alpha*Loss_{res}+Loss_{pred}
\end{align}
The determination of $\alpha$ considers two factors: the text and image losses should be comparable, and the changes in text and image losses during the training should be comparable. 
Such a multitask learning design does not require the introduction of additional labelled data and is highly relevant to the missing character prediction task.
\subsubsection{Curriculum Learning}

Curriculum learning involves initially focusing on simpler cases during the training and progressively ramping up the difficulty to tackle more intricate situations \citep{bengio2009}. In our task, images with smaller damaged areas are easier to identify. Consequently, we incrementally enlarge the damaged area with each training epoch. Assuming we want to simulate a damaged area with dimensions length $l$ and width $w$, curriculum learning is carried out over $k$ epochs. In the $j-th$ epoch, if $j < k$, both the $l$ and $w$ of this damaged area will be multiplied by $j/k$.

\section{Experiment}
\subsection{Baselines}
We employed single-modal methods as baselines, encompassing visual and textual modalities. In instances where the feature extraction models used in previous text restoration studies are outdated, we elevated them to frontier models and adapted them to ancient Chinese text restoration.

\begin{itemize}
    \item \textbf{Img}. When predicting damaged text using only visual features, it becomes similar to the handwritten Chinese character recognition (HCCR) task \citep{zhang2017online}, which has many mature models that can be adapted to text restoration. In our experiments, we use $ResNet\ 50$ \citep{he2016deep} to predict the character from the damaged image.

    \item\textbf{LM}. When predicting damaged text using only context features, like methods in \citet{ethan2020} and \citet{lazar-etal-2021-filling}, it reduces to a MaskLM task. We use the pre-trained masked language model $RoBERTa$ \citep{liu2019roberta} to predict the damaged character from context. This model, after pre-training on a large-scale dataset of ancient Chinese, possesses certain abilities in predicting masked characters.

    \item\textbf{LM ft}. Since pre-trained models serve multiple tasks, there may still be room for improvement in a single MaskLM task. We further fine-tuned the $RoBERTa$ model on our dataset.
\end{itemize}

\begin{table*}[ht]
\begin{center}
\begin{tabular}{|l|l|l|l|l|l|l|l|l|l|l|}
      \hline
      Model&Image&Text&LM ft&MT&CL&Acc&Hit 5&Hit 10&Hit 20&MRR\\
      \hline
      Img &+&&&&&66.00&79.43&83.33&86.55&72.18\\
      LM &&+&&&&36.06&56.18&64.10&71.03&45.56\\
      LM ft &&+&+&&&44.75&66.07&73.23&79.48&54.57\\
      \hline
      MRM &+&+&+&&&86.74&94.16&95.61&96.87&90.09\\
      MMRM &+&+&+&+&&87.34&94.60&96.16&97.29&90.61\\
      MMRM CL &+&+&+&+&+&\textbf{87.76}&\textbf{95.03}&\textbf{96.45}&\textbf{97.52}&\textbf{91.03}\\
      \hline
\end{tabular}
\caption{Results of simulation experiments. MT is for multitask learning, CL is for curriculum learning.}
\label{result_1}
\end{center}
\end{table*}

\subsection{Proposed Approach}
As for multimodal models, we initially fuse visual and text features to assess the multimodal approaches (MRM), and then introduce multitask learning (MMRM) and curriculum learning (MMRM CL) to further enhance the model.
\begin{itemize}
    \item\textbf{MRM} (Multimodal Restoring Model). We integrate both vision and text features to predict the damaged character in a multimodal manner. 

    \item\textbf{MMRM} (Multimodal Multitask Restoring Model). In addition to predicting the missing text token, this model performs multitask learning, allowing the model to generate the restored image of the damaged character. Unlike previous multitask methods such as \citet{assael2022restoring} and \citet{kang-etal-2021-restoring}, it requires no extra labelling data.

    \item\textbf{MMRM CL} (Multimodal Multitask Restoring Model with Curriculum Learning). During the training phase, curriculum learning is introduced to gradually increase the degree of damage, enabling the model to gain the restoration ability in a progressive manner.
\end{itemize}

\subsection{Metrics}
A text restoration task that relies solely on context can be evaluated based on the damaged amount and string length \citep{assael-etal-2019-restoring}. However, in simulated experiments involving visual signals, the masks undergo processes such as overlay and blur, making it hard to measure the degree of image damage in a standardized manner. Therefore, evaluation metrics in simulated experiments are to assess the performance of different model architectures under consistent simulated data. We chose three commonly used evaluation metrics in text restoration:
\begin{itemize}
    \item \textbf{Accuracy}. Accuracy stands for the average accuracy when the model predicts damaged characters.
    \item\textbf{Hits}. Hits represent the probability of the correct character being in the $top\ k$ candidates. Here we set $k$ to 5, 10, and 20.
    \item\textbf{MRR}. MRR stands for Mean Reciprocal Rank, which refers to the reciprocal of the rank of the correct character. 
\end{itemize}

\subsection{Settings}
In missing character sampling, to cater to a broader range of characters rather than concentrating on high-frequency ones, we assign weighted probabilities to each character. We first calculated the frequency of each character in the training dataset, the average frequency is $f_{avg}$. Assuming for character $Ci$, its frequency is represented as $f_i$, its sample weight $w_i$ is calculated as follows:
 \begin{equation}
     w_i=\sqrt{\frac{1}{max(f_i,f_{avg})}}
 \end{equation}
This setting helps to mitigate the influence of high-frequency characters. We used two sampling methods in the experiment: one is to mask one character to compare its performance with baseline models; the other is to randomly mask 1-5 characters to verify the model in the multiple missing characters scenarios. More technical details for simulating damaged images can be found in the Appendix
\footnote{Appendix: \url{https://github.com/CissyDuan/MMRM}.}.

In model building, we used the pre-trained $ResNet\ 50$ \footnote{The pre-trained $ResNet\ 50$ was downloaded from \url{https://download.pytorch.org/models/resnet50-19c8e357.pth}} and pre-trained $RoBERTa$ \footnote{The pre-trained $RoBERTa_{base}$ was downloaded from \url{https://github.com/Ethan-yt/guwenbert}, it has been pre-trained with Classical Chinese literature containing 1.7B characters.}. 
The damaged image simulation is conducted on 64x64 images. In the image decoder, the number of transposed convolution layers is 5. The loss weight $\alpha$ was set to 100. 

In training, the batch size is 256. The training lasted for 30 epochs, with curriculum learning applied in the first 10 epochs. The learning rate was set to 0.0001 and decayed to less than 1e-5. The optimizer is Adam \citep{kingma2014adam}. 

In metric calculation, all simulation results are the averages obtained after randomly sampling the damaged characters on the test set 30 times, ensuring statistical significance in the observed improvements.

\begin{table}
    \centering
    \resizebox{.99\linewidth}{!}{
    \begin{tabular}{|c|ccccc|}
        \hline
         Num&Acc  &Hit 5  &Hit 10  &Hit 20  &MRR \\
         \hline
          R&82.83	&91.57	&93.68	&95.30	&86.80\\
          1&87.27	&94.40	&95.99	&97.16	&90.50\\
          2&85.02	&92.95	&94.79	&96.28	&88.62\\
          3&82.67	&91.50	&93.59	&95.23	&86.67\\
          4&80.72	&90.01	&92.45	&94.41	&84.96\\
          5&78.76	&88.87	&91.44	&93.46	&83.38\\
         \hline
    \end{tabular}}
    \caption{Results for multiple missing characters. R stands for random missing 1-5 characters.}
    \label{tab:multiple}
\end{table}

\subsection{Simulation Results}
The results of simulation experiments are shown in Table \ref{result_1}. It can be observed that when using only context or image information, the prediction performance is not satisfactory. CNN-based models can achieve an accuracy of over 95\% in the task of undamaged handwritten Chinese character recognition (HCCR) \citep{zhang2017online, li2020}. However, in our simulation experiments, the $ResNet\ 50$ model, which performs well in the HCCR task, performed poorly in the damaged character recognition task ($Acc=66.00\%$). Likewise, although the $RoBERTa$ model, trained and fine-tuned for the MaskLM task, exhibited some predictive capability ($Acc=44.75\%$), it was insufficient for guiding real-world scenarios.

However, the introduction of the multimodal method (MRM) brought about a turning point, significantly improving prediction performance ($Acc=86.74\%$). This underscores the necessity of jointly leveraging text and image information, which was overlooked in previous ancient text restoration research. The introduction of multitask learning (MMRM) further enhanced restoration accuracy and provided restored images ($Acc=87.34\%$), and the curriculum learning approach (MMRM CL) also contributed a modest improvement to the model ($Acc=87.76\%$). These experimental results on simulated data validate the superior performance of our proposed MMRM architecture in ideograph restoration.

In multiple missing characters scenarios, shown in Table \ref{tab:multiple}, the results indicate that the model's performance slightly decreases but remains satisfactory. Moreover, in the single character damaged scenario, compared with specific training (Table \ref{result_1}, MMRM CL), there is only a very marginal decline in performance, suggesting that using one model is sufficient to address multiple types of characters missing.

\begin{figure}[t]
\begin{center}
\includegraphics[width=0.99\linewidth]{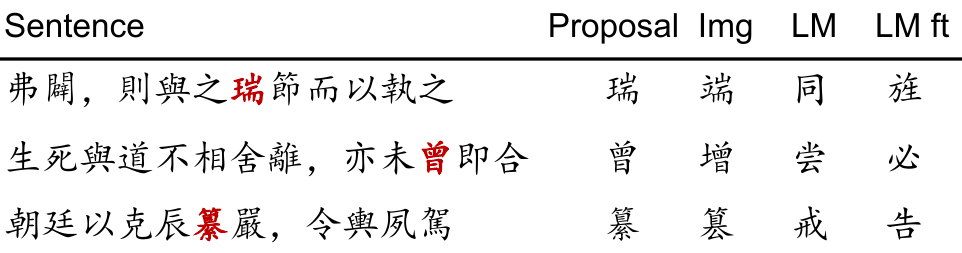} 
\caption{Cases from different models in simulation experiments}
\label{case study}
\end{center}
\end{figure}

We present cases from different models in simulation experiments in Figure \ref{case study}. Models that only use visual features will confuse text with similar shapes, and language models may suggest other characters that fit the context. 

It is important to note that these results are based on specific simulated data and are intended for comparing the performance of different models. This does not necessarily reflect the models' ability to perform text restoration in real-world scenarios.

\subsection{Real-world Scenario}
While we achieved commendable outcomes in the restoration of simulated data, it is still necessary to ascertain the model's continued efficacy within real-world scenarios. In this section, we conducted an empirical examination of the proposed model with a real historical artefact. 

The `Inscription of Sweet Spring in Jiucheng Palace (九成宫醴泉铭)' is a renowned calligraphic masterpiece during the Tang Dynasty (632). It was created by politician Wei Zheng and written by calligrapher Ouyang Xun and hailed as the `peerless regular script' of China. Over the passage of time, this inscription has endured significant damage. Humanities scholars have reached generally accepted restoration conclusions for its damaged characters through careful investigation of literature. This significance and the availability of expert suggestions make it an ideal candidate for evaluating our proposed method. 

\begin{figure}[h]
\begin{center}
\includegraphics[width=0.9\linewidth]{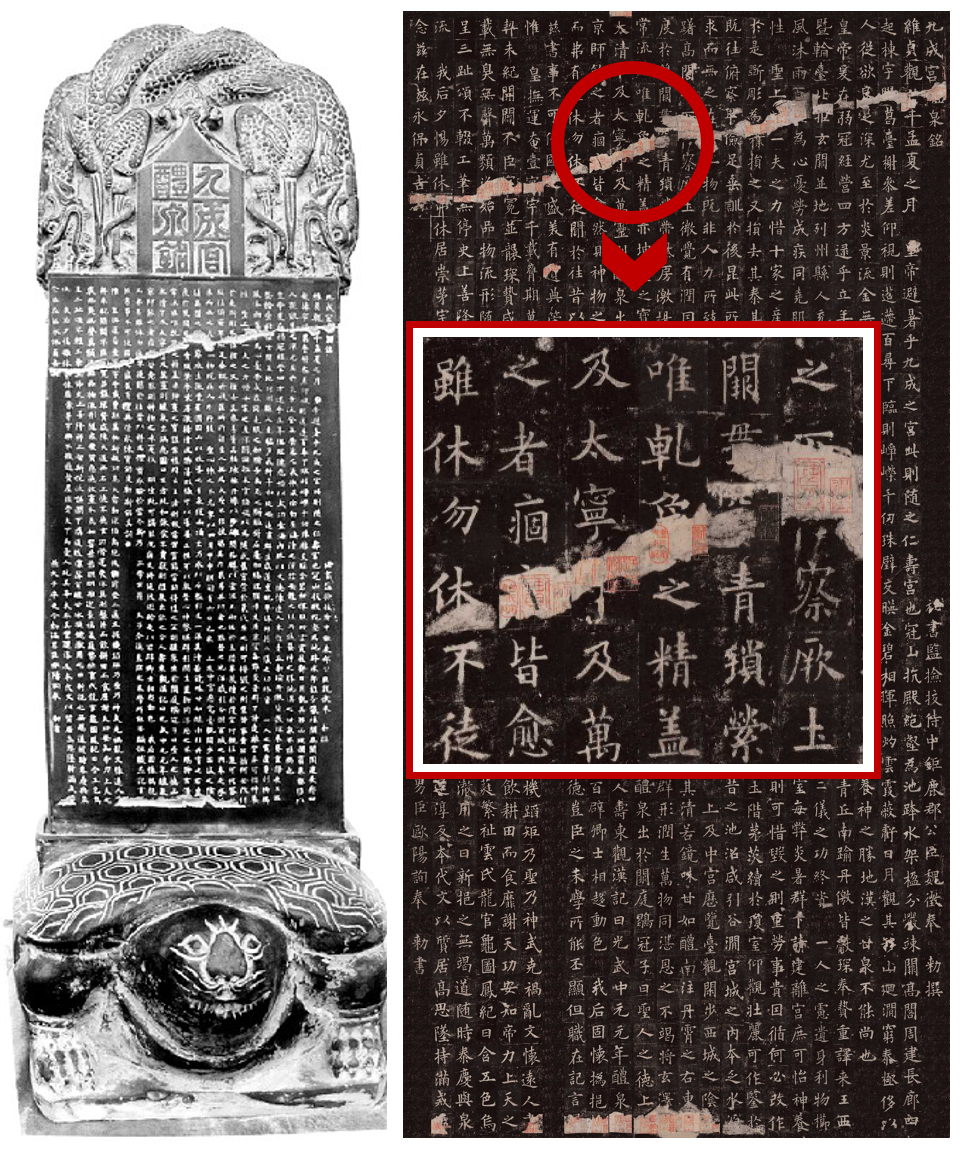} 
\caption{The full rubbing and photo of the `Inscription of Sweet Spring in Jiucheng Palace'}
\label{quan}
\end{center}
\end{figure}

The highest-quality extant rubbing of this inscription is the Li Qi edition, originating from the Song Dynasty (960 - 1279). We sourced a digital version of this edition from the internet and subjected it to restoration using the model developed in our simulation experiment. The full rubbing and photo of the inscription are shown in Figure \ref{quan}. Subsequently, we compared the outcomes generated by our model with the results suggested by experts to gauge the model's effectiveness in real-world scenarios.

\begin{figure*}[h]
\begin{center}
\includegraphics[width=0.8\textwidth]{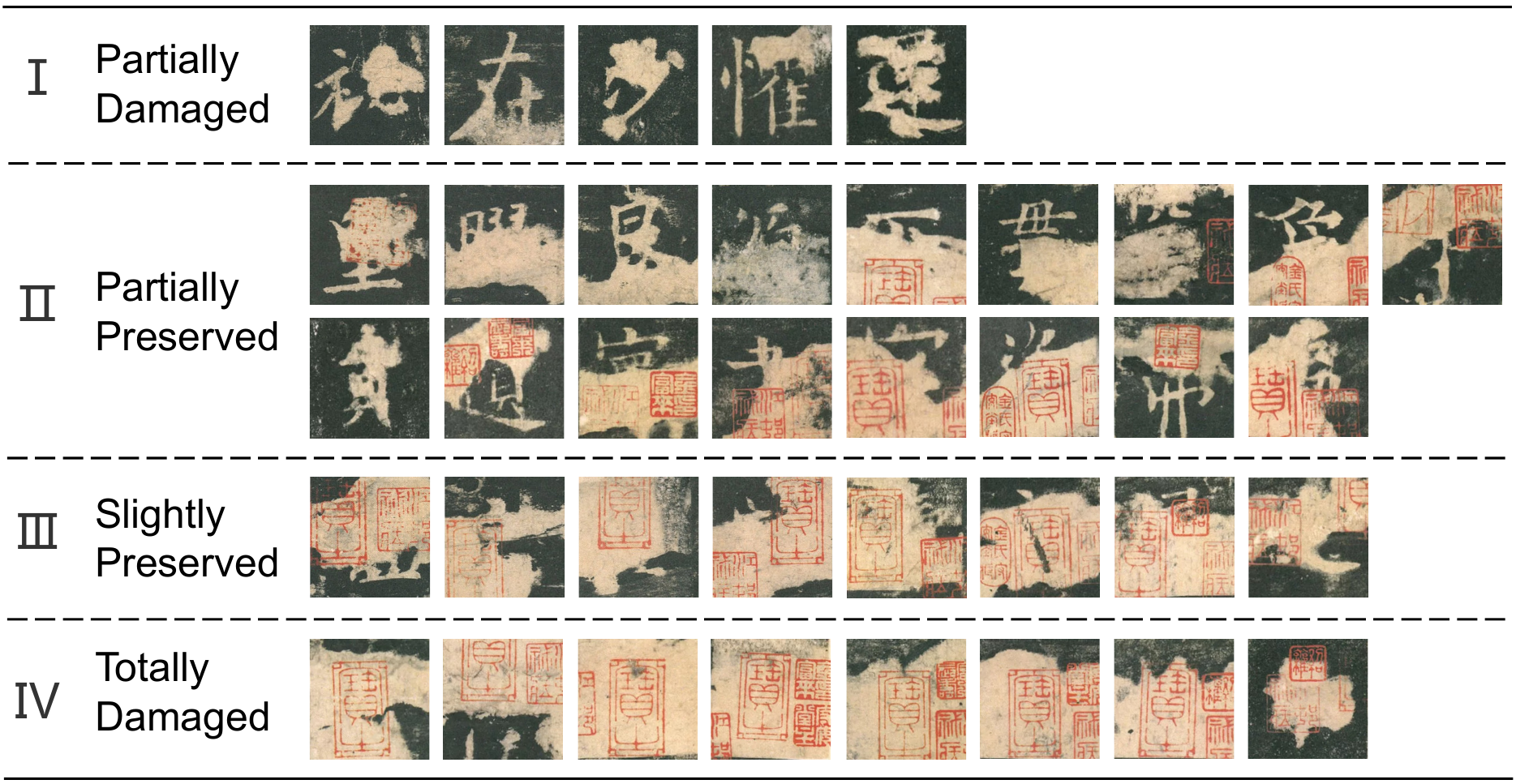} 
\caption{Four levels of damages from the `Inscription of Sweet Spring in Jiucheng Palace'.}
\label{real}
\end{center}
\end{figure*}

\begin{table*}[ht]
\begin{center}
\begin{tabular}{|l|r|r|r|r|r|r|r|r|r|r|}
      \hline
      Level&Number&Count&Accuracy&Hit 5&Hit 10&Hit 20&MRR\\
      \hline
      \uppercase\expandafter{\romannumeral1} &5&5&100.00&100.00&100.00&100.00&100.00\\
      \uppercase\expandafter{\romannumeral2} &17&11&64.70&82.35&82.35&94.11&71.17\\
      \uppercase\expandafter{\romannumeral3} &8&5&62.50&87.50&87.50&100.00&70.83\\
      \uppercase\expandafter{\romannumeral4} &8&0&0&25.00&37.50&50.00&10.06\\
      \hline
      Total &38&21&\textbf{55.26}&\textbf{73.68}&76.31&\textbf{86.84}&\textbf{62.28}\\
      \hline
      LM&38&11&28.94&68.42&\textbf{84.21}&86.84&46.57\\
       LM ft &38&13&34.21&73.68&78.95&84.21&50.32\\
       \hline
\end{tabular}
\caption{The results of real-world scenario experiments}
\label{result_2}
\end{center}
\end{table*}

We cropped images of 38 damaged characters from the rubbings. These images are categorized into four levels based on the degree of damage and are shown in Figure \ref{real}.
\begin{itemize}    
\item \textbf{\uppercase\expandafter{\romannumeral1}. Partially Damaged}: Some parts of the characters are damaged, but the overall shape is preserved. Experienced experts may be able to recognize the damaged characters from the remaining portions.
\item \textbf{\uppercase\expandafter{\romannumeral2}. Partially Preserved}: The characters in the images have a significant area of damage but still retain some key information, such as several strokes or radical components. While it narrows down the candidate's scope, variations in the damaged portion can point to different results.
\item \textbf{\uppercase\expandafter{\romannumeral3}. Slightly Preserved}: Most area of the character is damaged, and it is impossible to determine the missing characters from the image alone. However, minimal details, such as a single stroke, are preserved.
\item \textbf{\uppercase\expandafter{\romannumeral4}. Totally Damaged}: The images provide no useful information and are completely damaged. In this case, the restoration can only rely on contextual information.
\end{itemize}


In the preprocessing stage, we converted the images into grayscale images. Since there were several signature seals on the images, we manually masked these seals using similar colours to the neighbouring area. Afterwards, we resized the images to the dimensions required by the model. We used the MMRM CL model (Table \ref{result_1}, MMRM CL) to restore these texts, the results are shown in Table \ref{result_2}. The MMRM mentioned in the subsequent text all refer to MMRM CL. Due to the limited amount of real data, the MRR metric is more important since it can alleviate the sparsity to a certain extent.

As the results indicate, when compared to the language model that solely relies on textual data, our proposed method has demonstrated substantial improvements in accuracy and MRR within real-world scenarios. This underscores the effectiveness of our damaged image simulation and the MMRM framework.

Specifically, the MMRM model exhibits the capability to provide reasonable candidates for damage levels \uppercase\expandafter{\romannumeral1} to \uppercase\expandafter{\romannumeral3} ($MRR>70.00$). In the case of damage level \uppercase\expandafter{\romannumeral3}, where only minimal strokes are preserved, the model can still provide reliable recommendations ($MRR=70.83$). However, its effectiveness diminishes when confronted with damage level \uppercase\expandafter{\romannumeral4} ($MRR=10.06$), falling short of achieving the average MRR achieved by the language model ($MRR=50.32$). Consequently, in totally damaged cases, employing a language model for restoration assistance may be a more viable choice, while MMRM remains a suitable option when residual visual information is still present.


\begin{figure}[t]
\begin{center}
\includegraphics[width=0.99\linewidth]{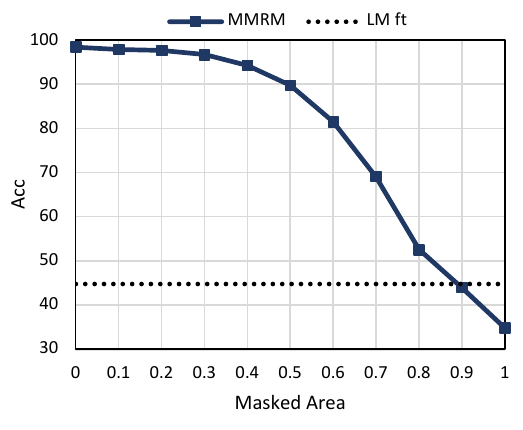} 

\caption{Relation between masked area and restoring accuracy}
\label{mask1}
\end{center}
\end{figure}



To further analyze the relationship between the degree of damage and model performance, we simulated damaged areas with different sizes separately. We added square-shaped damage of different sizes, using the side length ranging from 0 to 1.0 times the image size. The results are shown in Figure \ref{mask1}. 
It can be observed that when the side length of the damaged square is larger than 0.9 times the image's side length, the multimodal model is weaker than the fine-tuned language model. This indicates that when the damaged area is too large, it can no longer bring effective information but introduce extra noise. 

\begin{figure}[t]
\begin{center}
\includegraphics[width=0.99\linewidth]{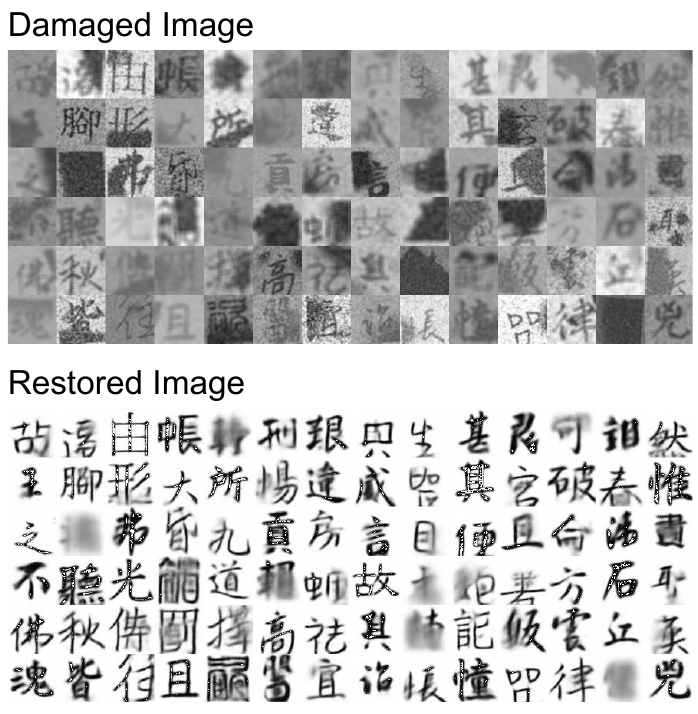} 
\caption{Restored images by MMRM in simulation experiment}
\label{restored pic}
\end{center}
\end{figure}

Additionally, we show the restored images from the simulation experiment in Figure \ref{restored pic}. Some severely damaged images can not achieve reasonable restoring results. In this case, it is a wiser choice to directly use the language model to assist text restoration.

\section{Limitations}
As the experimental results show, the proposed method still has limitations: it cannot give valuable results when facing large damaged areas; When multiple characters are damaged in the context, the model's performance inevitably drops. 
What's more, some ancient languages are endangered and do not have enough digital text and font resources to simulate the data, and the meanings of some characters may have not yet been deciphered, which poses further obstacles to their restoration.

In addition, there are challenges in generalizing to other ancient languages. Limited by our knowledge base, the experiments in this paper are conducted on Classical Chinese only. Scholars proficient in other ancient languages can try to adapt this method to other ancient texts.

\section{Conclusion and Future Work}

In this article, we propose a multimodal multitask model for ancient ideograph restoration, gaining a marked enhancement in simulation experiments and providing reasonable suggestions in real-world applications. This work represents a novel attempt that leverages the strengths of previous attempts in NLP and CV fields for restoring ancient textual artefacts, extending to situations bearing greater fidelity to real-world scenarios through multimodal methodologies.

There are many potential directions for future research: How to retrieve and utilize information from external databases to enhance text restoration? How to employ deep learning methods to aid in recognizing low-resource ancient ideographs, such as the ancient Chinese Oracle Bone Inscriptions from 3000 years ago? How to design an interactive tool for ancient text restoration, serving humanities scholars who lack the necessary programming skills? We look forward to the application of this work in both academic and industrial contexts.


\section{Acknowledgments}
This research is supported by the NSFC project “the Construction of the Knowledge Graph for the History of Chinese Confucianism” (Grant No. 72010107003).

\bibliographystyle{lrec-coling2024-natbib}
\bibliography{lrec-coling2024-example}

\bibliographystylelanguageresource{lrec-coling2024-natbib}
\bibliographylanguageresource{languageresource}

\end{CJK*}
\end{document}